\documentclass{article}
\usepackage{spconf, amsmath, graphicx}
\usepackage[style=ieee]{biblatex}
\usepackage{longtable}
\usepackage{float}
\usepackage{hyperref}
\usepackage{tabularx}
\usepackage{multirow}
\usepackage{booktabs}
\usepackage{multicol}
\usepackage{makecell}

\bibliography{isbi_for_arxiv_bib}
\defbibheading{bibliography}[]{}

\usepackage{enumitem}
\setlist{nosep, leftmargin=14pt}

\title{How and why does Deep Ensemble coupled with Transfer Learning increase performance in bipolar disorder and schizophrenia classification?}

\name{Sara Petiton$^1$, Antoine Grigis$^1$, Benoit Dufumier$^{1,2}$, Edouard Duchesnay$^1$ }
\address{$^1$ NeuroSpin, CEA Saclay, Université Paris-Saclay, France \\
         $^2$ EPFL, Lausanne, Switzerland}

\begin{document}
%
\maketitle
\begin{abstract}
Transfer learning (TL) and deep ensemble learning (DE) have recently been shown to outperform simple machine learning in classifying psychiatric disorders. However, there is still a lack of understanding as to why that is. This paper aims to understand how and why DE and TL reduce the variability of single-subject classification models in bipolar disorder (BD) and schizophrenia (SCZ). To this end, we investigated the training stability of TL and DE models. For the two classification tasks under consideration, we compared the results of multiple trainings with the same backbone but with different initializations. In this way, we take into account the epistemic uncertainty associated with the uncertainty in the estimation of the model parameters.
It has been shown that the performance of classifiers can be significantly improved by using TL with DE. Based on these results, we investigate i) how many models are needed to benefit from the performance improvement of DE when classifying BD and SCZ from healthy controls, and ii) how TL induces better generalization, with and without DE. In the first case, we show that DE reaches a plateau when 10 models are included in the ensemble. In the second case, we find that using a pre-trained model constrains TL models with the same pre-training to stay in the same basin of the loss function. This is not the case for DL models with randomly initialized weights. \footnote{{\scriptsize The scripts related to this study can be found at : \url{https://github.com/SaraMPetiton/DE_with_TL_study}}}

\end{abstract}


\section{Introduction}
\label{sec:intro}

Deep Learning (DL) has been shown to be an efficient way to classify medical images from MRI scanners \cite{DLMedicalImaging}. However, machine learning (ML) algorithms tend to perform as well, if not better, than DL when applied to some psychiatric disorders \cite{TheseBenoit}. 
It has recently been shown that deep ensemble (DE) and transfer learning (TL) paradigms outperform ML for single-subject prediction using 3D whole-brain anatomical MRIs \cite{TheseBenoit}. These newly proposed paradigms offer better performance on psychiatric classification tasks. Nevertheless, it isn't clear how TL enables this gain or to what extent DE improves predictions. 

In the case of psychiatric disorder classification, as for other medical prediction tasks, the reliability and robustness of predictions are very important. However, DL models whose weights have been randomly initialized (RI-DL) have multiple sources of variability \cite{variability}: aleatoric uncertainty, inherent to the data distribution, and epistemic uncertainty, also known as knowledge or model uncertainty \cite{uncertainty_DL}. In this study, we focus on the epistemic uncertainty associated with the random initialization of the model weights. Each time RI-DL models are trained, they may find a different set of weights, which in turn produce different predictions and different latent representations. A successful approach to reduce the variance of these models is to train multiple models and combine their predictions. This is known as ensemble learning \cite{DeepEnsemble}. It reduces prediction variance and can lead to better prediction performance. Unfortunately, training multiple models for DE can be time-consuming and computationally expensive. Therefore, finding a threshold for the number of models that need to be trained to see a significant performance improvement can save both time and resources.

On the other hand, TL can improve predictions by using a pre-trained model before fine-tuning previously acquired knowledge to the desired classification task \cite{TL}. Here, contrastive learning was used for pre-training using age as a weak supervision \cite{yaware}. The resulting predictions have been shown to outperform both ML and RI-DL. To illustrate and understand how TL works, a recent study proposed to compare models trained from different weights initializations using performance barrier curves \cite{transfer}. The authors compared TL and RI-DL models on natural images, drawings, and chest X-ray classification tasks. These comparisons were made by studying the effect of linear interpolation between the weights of any pair of models on a surrogate prediction task.

This paper aims to understand how and to what extent DE and TL outperform RI-DL for bipolar disorder (BD) and schizophrenia (SCZ) classification tasks. The proposed contributions are two-fold : 
\begin{itemize}
  \item[-] First, we show that 10 trained models are sufficient for the best performance improvement with DE for SCZ and BD. 
    
  \item[-] Secondly, we compare the loss landscapes of RI-DL and TL models using barrier curves. To the best of our knowledge, this is the first time that this method, proposed in \cite{transfer}, has been applied to whole-brain MRI datasets. We show that using TL with 3D MRIs for psychiatric classification tasks enables models to stay in the same basin of the loss landscape and presents a more robust approach than RI-DL. 
\end{itemize}

\section{Materials and Methods}
\label{sec:format}
\subsection{Datasets}
For SCZ classification, the datasets used are SCHIZCONNECT-VIP, CNP, PRAGUE, BSNIP, and CANDI, with 933 subjects used for training, 116 for validation, and 133 for testing. For BD classification, the datasets are BIOBD, BSNIP, CNP, and CANDI, with 832 subjects for training, 103 for validation, and 131 for testing. All splits are stratified on age, sex, site, and diagnosis, and the test sets include sites never seen during training to prevent overfitting on acquisition sites \cite{bias_test_sites}. CAT12 is used to compute voxel-based morphometry (VBM) gray matter maps \cite{cat12vbm}. These maps are used as input to the proposed TL and RI-DL models.

\subsection{Learning strategy}

In this study, we use a DenseNet 121 backbone\footnote{{\scriptsize \url{https://github.com/Duplums/SMLvsDL}}} from \cite{TheseBenoit}. This backbone, while limiting the number of parameters to be estimated, has been shown to give the best results on the psychiatric disorder classification tasks considered \cite{TheseBenoit}. The pre-trained model we used for TL was trained on healthy brains from the OpenBHB, HCP, OASIS 3, and ICBM datasets using a weakly-supervised contrastive learning method\footnote{{\scriptsize \url{https://github.com/Duplums/yAwareContrastiveLearning}}} \cite{yaware}. This pre-trained model uses the age-aware InfoNCE loss based on the hypothesis that capturing the biological variability in the healthy population
related to non-specific variables (in this case, age) with large datasets allows easier discovery of specific pathological variability. The pre-trained weights are then used as a starting point for the training of TL models with the same architecture as the RI-DL models.

During training, the models' learning rates decrease by a factor of $\gamma$ every $10$ epochs. This learning rate decay strategy aims to gradually take smaller steps during gradient descent as we get closer to a minimum of the loss function. We found that the optimal value of $\gamma$ should not be the same depending on the classification task. To tune this hyperparameter, we trained the considered TL and RI-DL models for 200 epochs with $\gamma$ equal to 0.2, 0.4, 0.6, and 0.8. The ROC-AUC metric, which is well suited for the binary classification tasks considered (Healthy Controls vs. BD and Healthy Controls vs. SCZ), is used to evaluate model performance.

\subsection{Deep ensemble learning}

For each sample, we grouped $T$ models (either TL or RI-DL) and computed the average of their predicted labels, viewing it as a distribution estimation of $p(y|x, \mathcal{D})$, with $x$ the input image, $\mathcal{D}$ the training set, and $y$ the clinical status:

\begin{equation}
\hat{p}^{T}(y|x, \mathcal{D}) = \frac{1}{T}\sum_{t=1}^{T}p(y|x,\theta^{(t)}) \approx \frac{1}{T}\sum_{t=1}^{T} \hat{y}_{\theta_{t}} = \hat{y}^{T-DE}
\label{eq:DE}
\end{equation}

where T is the number of trained models, $\theta$  the model's weights, and $\hat{y}^{T-DE}$ corresponds to the predicted labels from DE averaging. This averaging minimizes the epistemic uncertainty of the models \cite{AverageEnsemble_to_reduce_uncertainty}. It has already been shown in the literature that the use of DE with TL leads to better results compared with RI-DL or TL alone \cite{TheseBenoit}. Here, we investigate how many models are needed to benefit from the performance improvement brought by DE and how the number of models influences performance variability. 

From $N=90$ trained TL models or $N=90$ trained RI-DL models $\{f_{\theta_1}, ..., f_{\theta_N}\}$, we get individual predictions $\hat{y}_{\theta_{i}}$, where $i \in [\![1, N]\!]$. 

Then, we draw with replacement (bootstrap) $P$ subsets of $T$ models with $T \in \{2, 5, 10, 15, 20, 30, 40, 50, 60\}$, from which we compute an ensemble score $\hat{y}^{T\text{-DE}}_{p}$ using Eq. \ref{eq:DE}, where $p \in [\![1, P]\!]$. There is no significant computational overhead in using a large value of $P$, since we are only bootstrapping the predictions. After testing several values of $P$, we chose $P = 10^{5}$.

\subsection{Linear interpolation of TL and RI-DL models}
To understand why TL outperforms RI-DL for single subject classification of SCZ and BD, we applied the linear interpolation method proposed in \cite{transfer}, which linearly interpolates pairs of TL and RI-DL model weights to look for barriers in the loss landscape.
The choice of using linear interpolation as a way to study the flatness of the loss landscape near a solution was discussed in \cite{transfer}. In \cite{lininterptheory1}, \cite{low_loss_path}, and \cite{low_loss_path2}, authors demonstrate that two minima in any DL model loss landscape can always be connected by a non-linear path maintaining a low loss. By contrast, finding whether the linear interpolation path between two DL models maintains a low loss or not enables us to decipher whether our trained models lie in the same local minimum of the loss function. 

We performed linear interpolations between TL and RI-DL models.
The weights of the interpolated models along the linear interpolation path are calculated as follows:
\begin{equation}
\theta_{\lambda}=(1-\lambda)\theta_1+\lambda \theta_2
\label{eq:LI}
\end{equation}

where $\lambda\in [0,1]$ is the linear interpolation coefficient, $\theta_1$ the weights of the first model, and $\theta_2$ the weights of the second. In practice, we used 30 values of $\lambda$, uniformly distributed between 0 and 1. 

Given a pair of trained models, we examine the behavior of the models obtained along such a linear interpolation path. If the chosen performance metrics remain good along this path, then no performance barrier is met, meaning that the two input models rest in the same basin of the loss landscape. 
Conversely, if the performance metric drops or is highly irregular along the path, it means that a performance barrier was encountered and that the two input models do not lie in the same basin of the loss landscape. For our classification tasks, this barrier will materialize as a decrease in the chosen performance metric, i.e., the ROC-AUC.

The experimental setup compares two RI-DL and two TL models initialized with the same pre-trained weights. Interestingly, in \cite{transfer}, the authors also looked at the linear interpolation between models at their last training epoch and at the epoch at which they perform best. We replicated this experiment to see if the TL models converge faster than the RI-DL models. We linearly interpolated TL and RI-DL models at their last training epoch (we trained them for 200 epochs) and at their best-performing epoch. We will refer to the former models as RI-DL and TL, and to the latter as RI-DL$^{*}$ and TL$^{*}$. Finally, we propose to study the following scenarios for the two classification tasks considered: TL to TL, RI-DL to RI-DL, TL to TL*, and RI-DL to RI-DL*.

\section{Results}
\label{sec:format}
\subsection{DE performance improvement reaches a plateau}

\begin{table}[tb]
  \scriptsize
  \setlength{\tabcolsep}{3pt}
  \begin{tabular}{c c c c c c}
    \toprule
    \multirow{2}{*}{\textbf{Task}} & \multirow{2}{*}{\textbf{Strategy}} & \multirow{2}{*}{\textbf{Baseline}} & \multicolumn{3}{c}{\textbf{Deep Ensemble}} \\
    \cmidrule(lr){4-6}  
    & & & \textbf{T=3} & \textbf{T=10} & \textbf{T=40} \\
    \midrule
    \multirow{2}{*}{\makecell{BD\\classification $\uparrow$}} & TL & $74.68_{\pm 1.96}$ & $76.24_{\pm 1.26}$ & $\textbf{77.06}_{\pm 0.74}$ & $77.53_{\pm 0.41}$ \\
    & RI-DL & $71.19_{\pm 2.8}$ &  $73.36_{\pm 1.76}$ & $74.55_{\pm 1.04}$ & $75.07_{\pm 0.55}$ \\
    \midrule
    \multirow{2}{*}{\makecell{SCZ\\classification $\uparrow$}} & TL & $72.76_{\pm 1.65}$ & $73.56_{\pm 1.05}$ & $\textbf{73.94}_{\pm 0.63}$ & $74.12_{\pm 0.35}$ \\ 
    & RI-DL & $72.51_{\pm 2.1}$ & $73.76_{\pm 1.35}$ & $74.16_{\pm 0.79}$ & $74.3_{\pm 0.42}$ \\
    \bottomrule
  \end{tabular}
    \caption{ROC-AUC (in \%) with standard deviations for both BD and SCZ classification tasks. Randomly initialized DL (RI-DL) models are compared with transfer learning (TL) models. In both cases, we evaluate the benefit of using deep ensemble (DE) learning. In our setting, "Baseline" corresponds to "no-DE", and T is defined in Eq. \ref{eq:DE}.}
\label{tab:ROCAUC}
\end{table}

\begin{figure}[tb]
\begin{minipage}[b]{1.0\linewidth}
  \centering
  \centerline{\includegraphics[width=9.0cm]{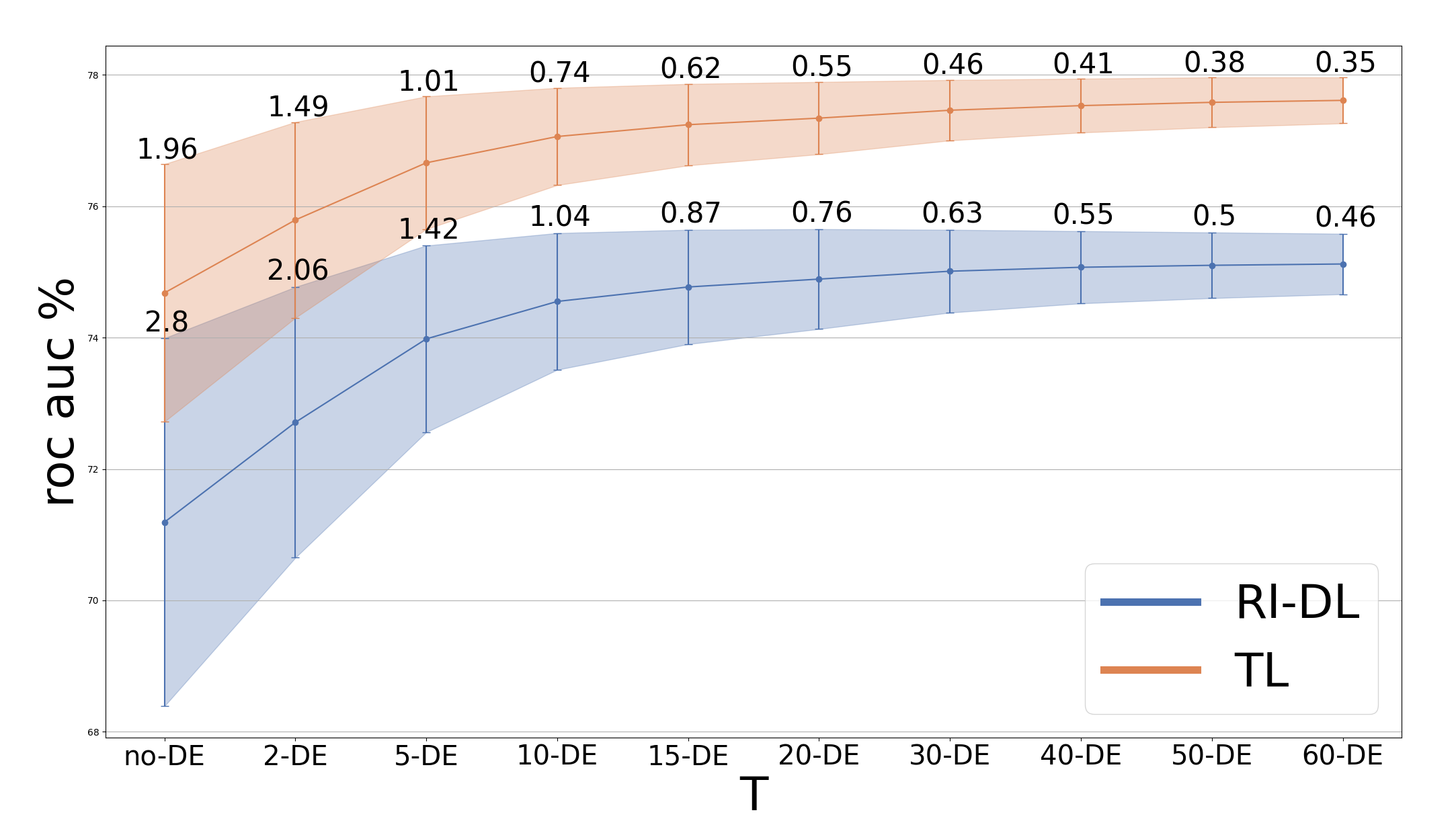}}
  \end{minipage}
\caption{Learning curves obtained by monitoring the ROC-AUC performance of BD classification as a function of the number of models T considered in the deep ensemble (DE) strategy. The obtained standard deviations are shown directly in the figure for each $T$-DE value examined on the x-axis. The "x=no-DE" configurations correspond to the means and standard deviations of the 90 trained models without DE.}
\label{fig:DE-ROC}

\end{figure}

To improve SCZ and BD classification using DE, we searched for an optimal number of models to train. From two sets of 90 RI-DL models and 90 TL models, we investigated the performance of DE learning as a function of the number of models $T$ considered (see Eq. \ref{eq:DE}). The results are shown in the Table \ref{tab:ROCAUC} and Figure \ref{fig:DE-ROC}. Overall, we found that DE performance with TL reaches a plateau when using $T=10$ models (10-DE) for both BD and SCZ classification tasks and that the most robust and accurate predictions were obtained by using TL with DE.
\\ 
\indent More specifically, from Table \ref{tab:ROCAUC} and for the BD classification task, the mean gain in ROC-AUC from TL with no-DE to TL with 40-DE is 2.85\%. Similarly, the gain in ROC-AUC from TL with no-DE to TL with 10-DE is 2.38\%. Therefore, the gain from using 40 instead of 10 models for DE is only 0.47\%. 
We witnessed similar trends when using RI-DL models. The ROC-AUC increases by 3.88\% with 40-DE compared with no-DE, and by 3.36\% with 10-DE.
For the SCZ classification task, the gain from TL with no-DE to TL with 40-DE is 1.36\%, compared with 1.18\% for TL with 10-DE.
For the RI-DL models, the ROC-AUC gain from no-DE is 1.79\% for 40-DE and 1.65\% for 10-DE.
We can see that the improvements of the ROC-AUC from no-DE to 10-DE and from no-DE to 40-DE are very similar. Looking at the learning curve in Figure \ref{fig:DE-ROC}, we confirm this observation. We can see that the ROC-AUC starts to reach a plateau after 10-DE.
\\ 
\indent From Table \ref{tab:ROCAUC}, we can also see that TL with DE outperforms RI-DL with DE only in the case of BD. For SCZ classification, the TL and RI-DL models have very close ROC-AUC performances. TL with 40-DE gives 2.46\% higher ROC-AUC values than RI-DL with 40-DE for BD. For SCZ, TL with 40-DE gives 0.18\% lower ROC-AUC values than RI-DL with 40-DE. 
\\ 
\indent In all cases, we see that as the number of models used in DE increases, the ROC-AUC increases, and the associated standard deviation decreases (see Figure \ref{fig:DE-ROC} for the BD classification task). Note that the standard deviation is also always lower with TL models. The most robust predictions are therefore obtained by using TL with DE.

\subsection{Transfer learning minimizes variability of trained models}

\begin{figure}[tb]
\begin{minipage}[b]{1.0\linewidth}
  \centering
  \centerline{\includegraphics[width=6.5cm]{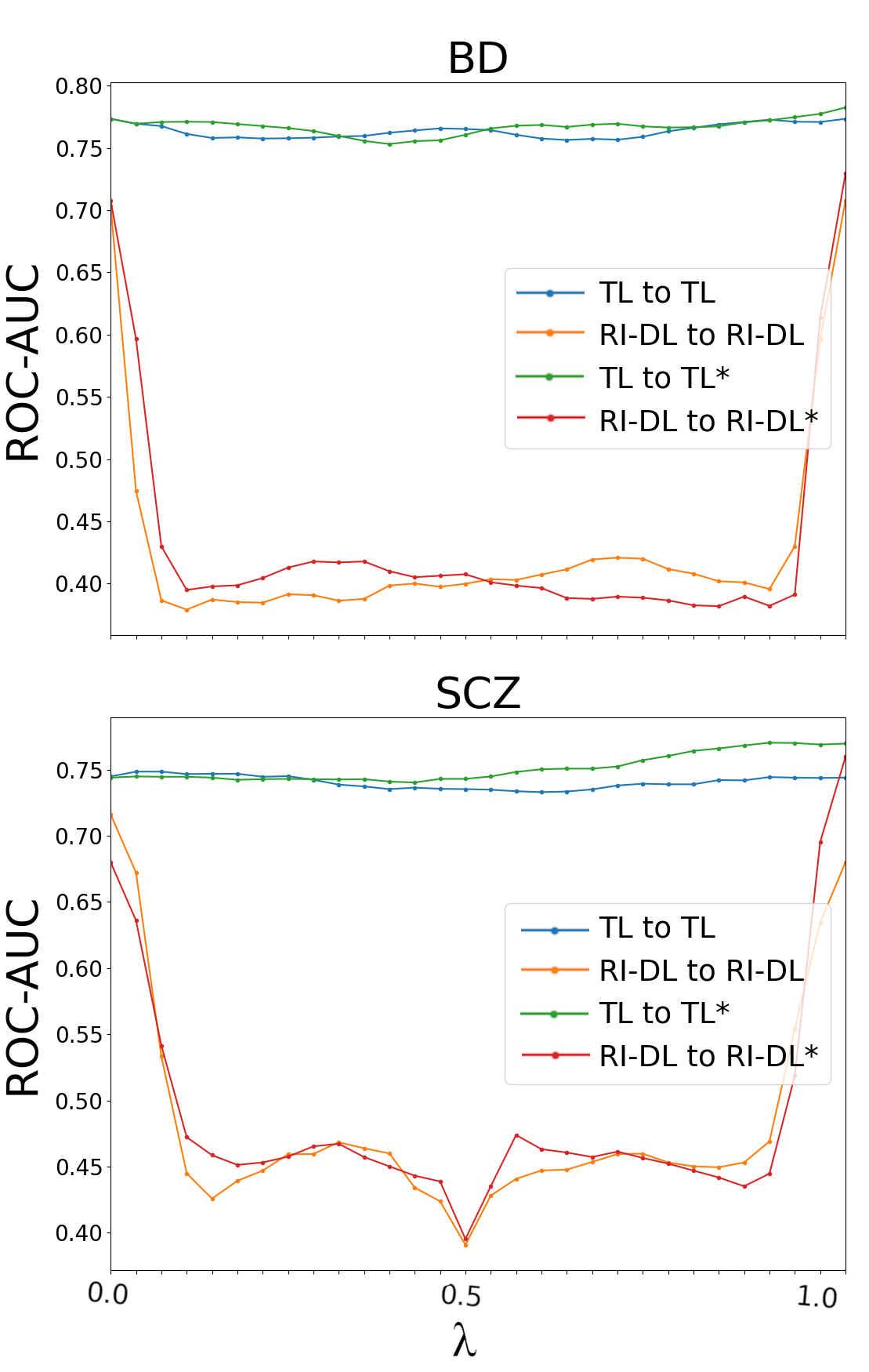}}
  \end{minipage}
\caption{Linear interpolation between RI-DL and TL models at the last and best training epochs on both BD and SCZ datasets. $\lambda\in [0,1]$ is the linear interpolation coefficient (see Eq. \ref{eq:LI}).}
\label{fig:res}
\end{figure}

In Figure \ref{fig:res}, we have plotted the evolution of the ROC-AUC performance metric along the linear interpolation path between two selected models in the two classification tasks considered (BD and SCZ).
In both cases, the ROC-AUC remains high and resembles an almost straight line when linearly interpolating between the weights of two models trained with TL (blue and green curves in Figure \ref{fig:res}). This means that the TL and TL* models remain in the same basin of the loss landscape, as they do not encounter a barrier that would cause the ROC-AUC to drop along the x-axis.
However, the ROC-AUC along the x-axis when two RI-DL models are interpolated (orange and red curves in Figure \ref{fig:res}) decreases when $\lambda$ (the linear interpolation coefficient) is close to zero, and increases again when $\lambda$ is close to 1. This means that the considered input RI-DL models encounter a barrier in the loss landscape and thus fail to complete their training in the same loss basin.
This shows that the TL models do not tend to fall into different local minima of loss during the fine-tuning process. The TL models are, therefore, more reliable than the RI-DL models, as they predict results with higher consistency for the two classification tasks considered.

The interpolation of TL to TL* models (the green curves in Figure \ref{fig:res}) shows that TL models remain in the same basin of the loss landscape from their best-performing epoch to their last training epoch. This is not the case for RI-DL models (the red curves in Figure \ref{fig:res}), where a decrease in the performance metric along the x-axis is observed for both BD and SCZ classification tasks. 
This means that once TL models reach their best ROC-AUC performance metric, they will remain in the same loss basin for the rest of the training. Therefore, their performance capability will remain the same once their best-performing epoch has been reached. Therefore, the TL models we studied not only perform better and with less variability than the RI-DL models, but they also require fewer training epochs.

\section{Conclusion}

In this paper, we explore how TL and DE can improve the performance of single-subject classification of BD and SCZ.
We show how both techniques can reduce model variability. 
In particular, the variability reduction achieved by DE learning depends on how many trained model predictions are averaged. In our two applications, we have shown that ten models are sufficient for this averaging to be beneficial both in terms of performance and variability reduction, as well as model robustness.
Furthermore, we show that TL maintains BD and SCZ classification models in the same basin of the loss landscape. Indeed, TL prevents the trained models from moving to different basins during fine-tuning. As a result, these models produce similar predictions. Compared with RI-DL, TL provides better, more robust predictions, and requires fewer training epochs. 
\\ 
\indent Overall, this work sheds light on the underlying mechanisms of performance improvements when using TL and DE in psychiatric disorder classification. We have shown that (i) 10 trained models are sufficient to achieve excellent and robust predictions when using DE and (ii) that TL models using 3D whole-brain MRI data provide coherent results by staying in the same basin of the loss landscape.
\\ 
\indent Further work could investigate why TL using age-aware contrastive learning \cite{yaware} as pre-training benefits some psychiatric disorders more than others in comparison with RI-DL.  
In \cite{TheseBenoit}, it is suggested that TL might not perform as well on the SCZ classification task as BD due to a simplicity bias \cite{simplicity_bias} hindering model generalizability. Indeed, SCZ subjects have been shown to display stronger deviations from healthy controls than BD subjects \cite{heterogeneitySCZ_BD}, making the classification of SCZ against healthy controls an easier task.
Moreover, some studies \cite{brain_aging_scz} \cite{brain_aging_scz_bd} have shown that SCZ is associated with accelerated brain aging (much more so than BD), indicating that there might be an overlap between the pre-training model and the classification task. 

\section{Acknowledgments}

This work was funded by Big2Small (ANR-19-CHIA-0010-01), RHU-PsyCARE (ANR-18-RHUS-0014), and R-LiNK (H2020-SC1-2017, 754907).

\section{Compliance with Ethical Standards}

This research study was conducted retrospectively using human subject data under the local ethics policy.

\section{References}
\label{sec:ref}

\printbibliography

\end{document}